\pdfoutput=1

\documentclass[11pt]{article}

\usepackage[preprint]{acl}
\usepackage{subcaption}
\usepackage{tcolorbox}

\usepackage{times}
\usepackage{latexsym}
\usepackage{float}

\usepackage{hyperref}
\usepackage{url}
\usepackage{amsmath}
\usepackage{amsthm,amsfonts}
\usepackage{xspace}
\usepackage{algorithm}
\usepackage{algorithmicx}
\usepackage{algpseudocode}
\usepackage{subcaption}
\usepackage{graphicx}
\usepackage{enumitem}
\usepackage{booktabs}
\usepackage{multirow}
\usepackage{bm}
\usepackage{xspace}

\usepackage{etoolbox}

\usepackage{tcolorbox}
\tcbuselibrary{listingsutf8}

\newtcolorbox{examplebox}[1][]{
  colback=blue!5!white, colframe=blue!75!black, 
  fonttitle=\bfseries, title=#1, 
  boxrule=0.5mm, sharp corners, 
  width=\linewidth, 
  fontupper=\small,
  listing options={basicstyle=\ttfamily\small, breaklines=true},
}

\makeatletter
\renewcommand{\alglinenumber}[1]{\fontsize{8}{8}\selectfont#1}
\newcommand\ALG@numberline{%
    \stepcounter{ALG@line}%
    \ifnum\value{ALG@rem}>0
        \stepcounter{ALG@rem}%
    \else
        \setcounter{ALG@rem}{9}%
        \alglinenumber{\arabic{ALG@line}}\ %
    \fi
}
\patchcmd{\ALG@step}{\relax}{\ALG@numberline}{}{}
\makeatother

\usepackage[T1]{fontenc}

\usepackage[utf8]{inputenc}

\usepackage{microtype}

\usepackage{inconsolata}

\usepackage{graphicx}

\newcommand{\E}{\mathbb{E}}
\newcommand{\method}{\textsc{Osca}\xspace}

\title{Scaling LLM Inference with Optimized Sample Compute Allocation}

\author{
 \textbf{Kexun Zhang\textsuperscript{1}\thanks{Equal contribution. Correspondence to \texttt{kexun@cmu.edu}.}},
 \textbf{Shang Zhou\textsuperscript{2}$^*$},
 \textbf{Danqing Wang\textsuperscript{1}},
 \textbf{William Yang Wang\textsuperscript{3}},
 \textbf{Lei Li\textsuperscript{1}}
\\
\\
 \textsuperscript{1}Carnegie Mellon University,
 \textsuperscript{2}UC San Diego,
 \textsuperscript{3}UC Santa Barbara
}

\begin{document}
 \maketitle
\begin{abstract}
Sampling is a basic operation in many inference-time algorithms of large language models (LLMs).
To scale up inference efficiently with a limited compute, it is crucial to find an optimal allocation for sample compute budgets:
Which sampling configurations (model, temperature, language, etc.) do we use?
How many samples do we generate in each configuration?
We formulate these choices as a learning problem and propose \method, an algorithm that \underline{O}ptimizes \underline{S}ample \underline{C}ompute \underline{A}llocation by finding an optimal mix of different inference configurations.
Our experiments show that with our learned mixed allocation, we can achieve accuracy better than the best single configuration with 128x less compute on code generation and 25x less compute on 4 reasoning tasks.
\method is also shown to be effective in agentic workflows beyond single-turn tasks, achieving a better accuracy on SWE-Bench with 3x less compute than the default configuration.
Our code and generations are released at \url{https://github.com/LeiLiLab/OSCA}.
\end{abstract}

\section{Introduction}

Large language models (LLMs) solve more problems with more inference compute.
Different ways of scaling up LLM inference include sampling \citep{chen2021evaluating}, self-consistency \citep{wangself}, tree search \citep{yao2024tree}, and multi-agent systems \citep{duimproving}, etc.
Among these, sampling is the most basic and serves as an atomic operation needed in all other more complicated methods.
Therefore, it is crucial to do it well.

\begin{figure}[th!]
      \centering
	   \begin{subfigure}{0.9\linewidth}
		\includegraphics[width=\linewidth]{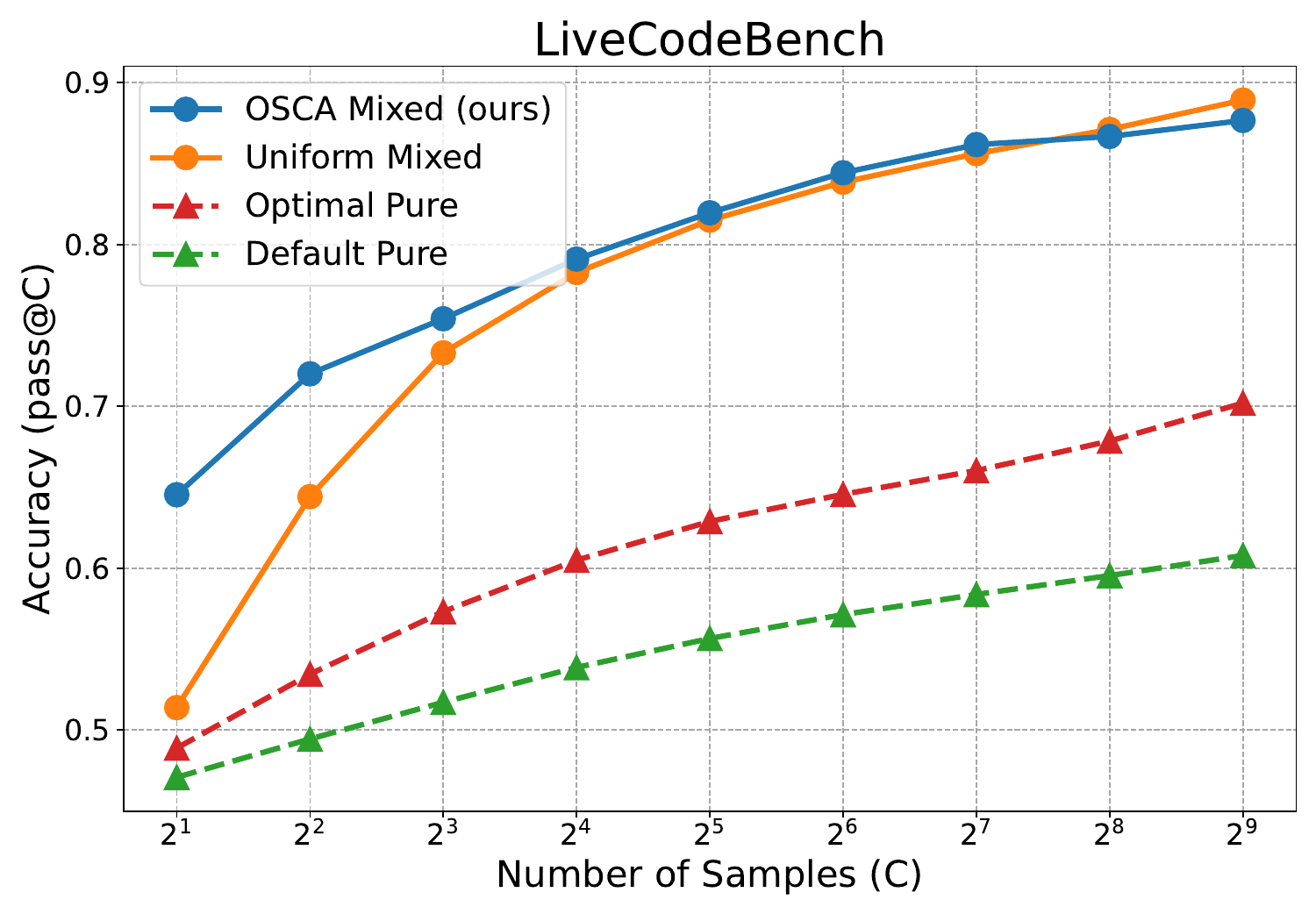}
	\end{subfigure}
 \hfill
	   \begin{subfigure}{0.9\linewidth}
		\includegraphics[width=\linewidth]{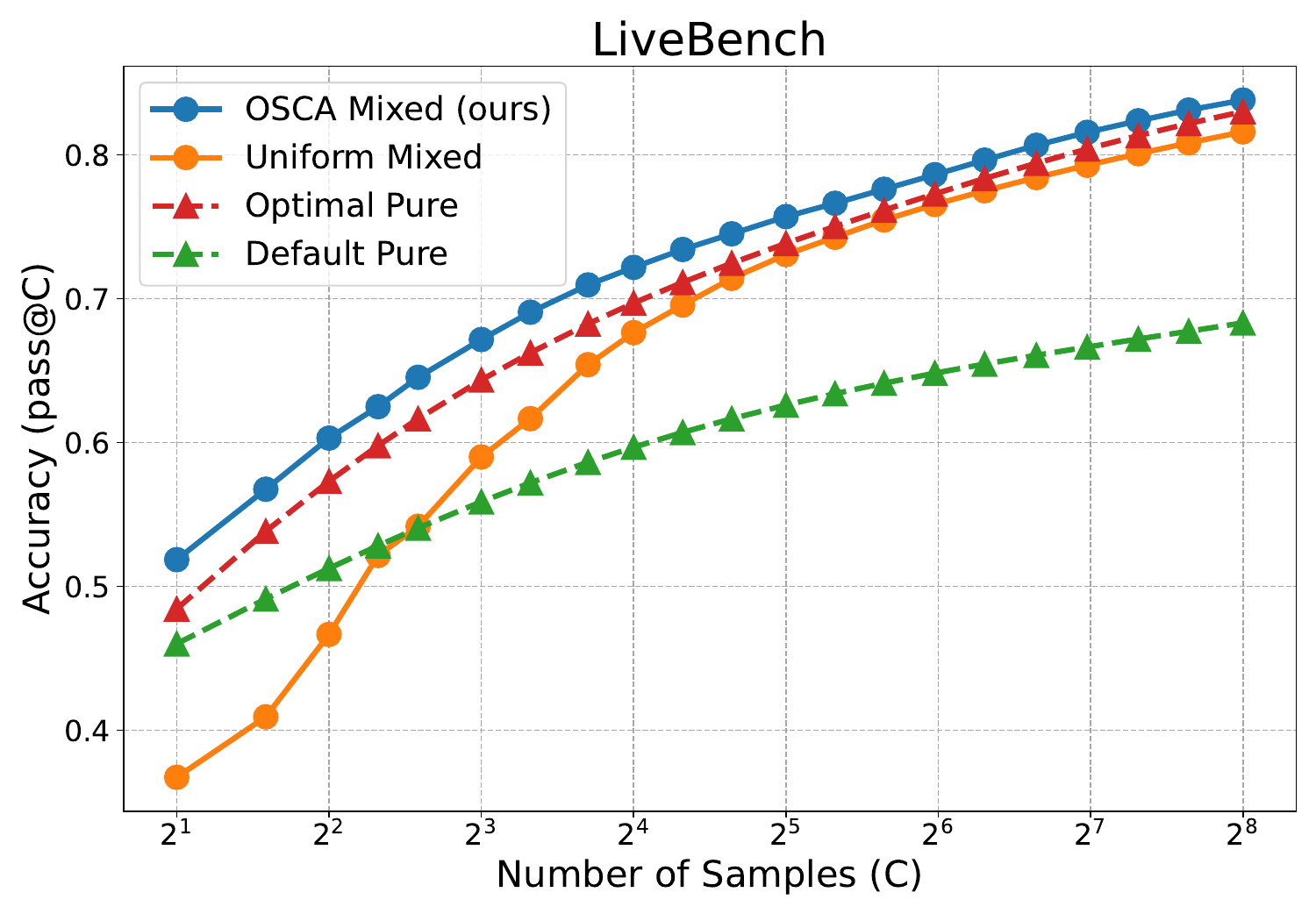}
		\end{subfigure}
 \hfill
 \begin{subfigure}{0.9\linewidth}
		\includegraphics[width=\linewidth]{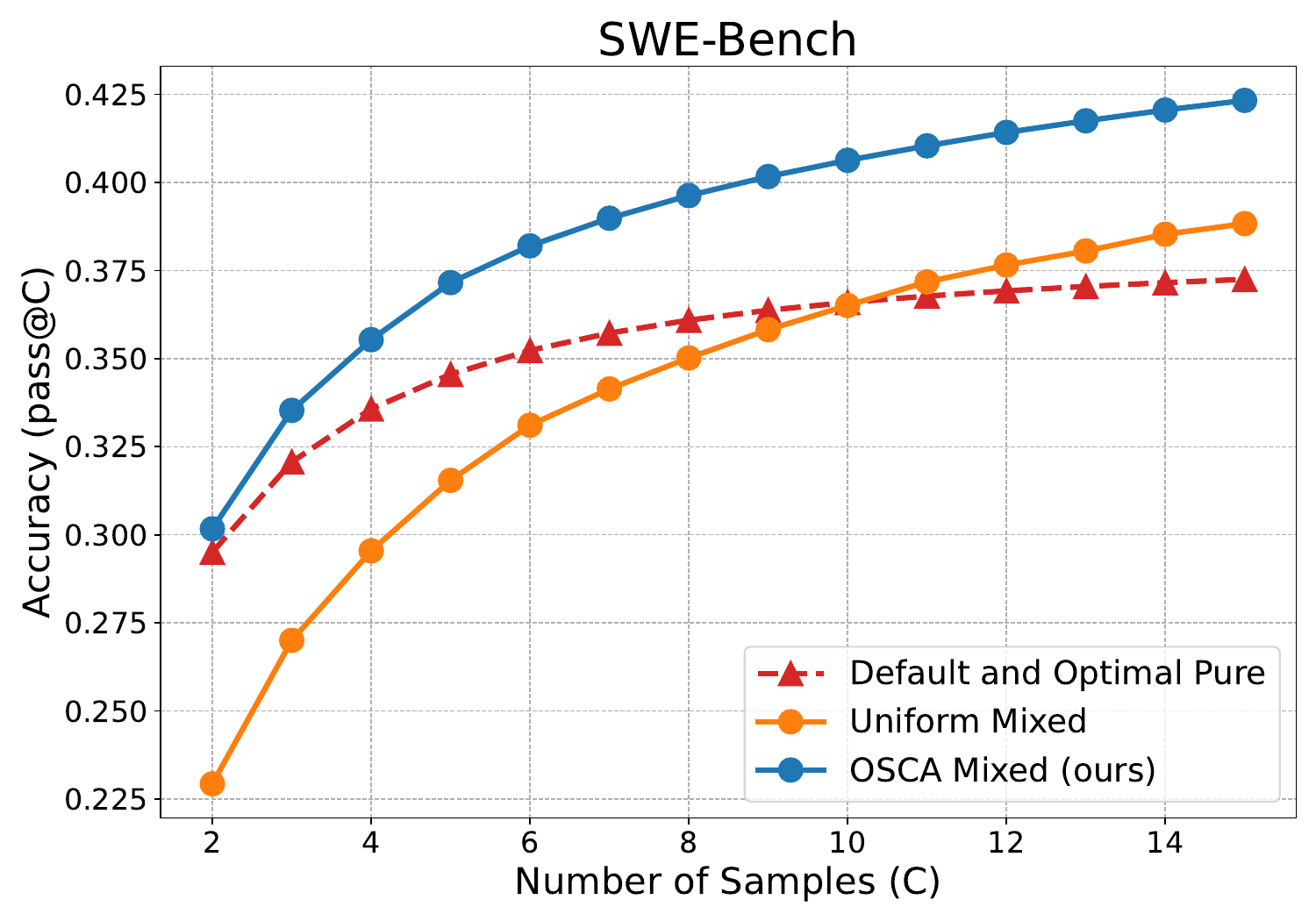}
		\end{subfigure}
 \caption{On 2 single-turn benchmarks and 1 agentic benchmark with a total of 6 tasks, our optimized allocations of sample compute are better than both optimal pure allocations and uniform allocations in most cases, especially when the compute budget is small.}
\label{fig:fig1}
 \vspace{-16pt}
\end{figure}

Previous studies \citep{wang2023cost} have investigated how to find the optimal sampling configuration, such as the best temperature, model, and prompt. While these methods are effective, they miss one key fact: \textbf{Not all problems require the same optimal sampling configuration.}
Some problems are easier solved with higher temperatures while others with lower temperatures \citep{li2022competition}. In this case, the best way to use a limited compute budget is not to choose between high or low temperatures, but to split the budget between both. This highlights the need for a \textit{mixed} allocation of the sample budget instead of a \textit{pure} allocation that uses a single configuration.

As shown in \autoref{fig:fig1}, just by uniformly allocating the compute budget over all possible inference configurations (dubbed ``uniform mixed''), LLMs' accuracy can already surpass the optimal pure allocation on LiveCodeBench.
Uniform mixed allocation gets 64\% accuracy with 4 samples, while optimal pure needs 16x more samples to get a similar accuracy.
Since uniform allocation is only one of the exponentially many allocations in the search space, it is natural to ask:
\textit{\textbf{How do we find the optimal sample budget allocation for LLM inference?}}

We formulate this as a learning problem: given a set of different inference configurations, a training problem set, and a compute budget, we need to distribute the budget over different configurations such that the expected accuracy is maximized.
To solve this problem, we propose a hill-climbing algorithm and demonstrate its effectiveness with both theoretical justification and experiments.
As shown in \autoref{fig:fig1}, the learned allocation from \method is significantly better than the optimal pure allocation, especially when the sample size is small.

To provide further insights for future adopters of \method, we conduct ablation studies on the effectiveness of different hyperparameters in the mix, the number of problems needed in the training set, and scenarios where mixed allocations do not offer significant improvements. Moreover, we show on SWE-Bench, a benchmark for LLM agents, that replacing pure sampling with \method's learned allocation in just one step boosts the entire workflow's performance. This demonstrates that a mixed sampling strategy not only improves single-turn tasks but also enhances complex inference algorithms, leading to better reasoning and decision-making.

Our contributions are the following:
\begin{itemize}[noitemsep, leftmargin=1em, topsep=1pt]
    \item We highlight the need for mixed allocation of LLM sample compute and formulate it as a learning problem.
    \item We propose an effective algorithm \method for optimizing sample compute allocations and demonstrate its effectiveness on 3 benchmarks consisting of 6 LLM tasks.
    \item We provide detailed analyses of when and why mixed allocations work, as well as their role in more complicated inference time algorithms.
\end{itemize}

\section{Related Work}

\textbf{Inference Time Algorithms.}
Following the taxonomy of \citet{welleck2024decoding} on inference-time algorithms, \textit{chained meta-generators} run multiple LLM calls sequentially and use the output sample from each call as the input to the next one \citep{dohan2022language,schlag2023large}.
\textit{Parallel meta-generators} samples multiple candidates for a problem and selects the best candidate %
\citep{wangself,chen2022codet,jiang2023LLMBlenderEnsemblingLarge,zhang2023algo,huang2023enhancing}.
\textit{Step-level search methods} regards problem-solving as a multi-step process and sample candidate next steps at each intermediate state, using algorithms like tree search \citep{yao2024tree}, graph search, and Monte-Carlo Tree Search \citep{lample2022hypertree,tian2024selfimprovementllmsimaginationsearching,chi2024thoughtsculpt}. %
\textit{Refinement-based} methods samples candidate solutions sequentially, relying on some feedback to revise the next candidate \citep{madaan2024self,shinn2024reflexion}.
Although these algorithms scale up inference differently, they all need LLM sampling as a basic operation.

\noindent \textbf{Scaling Inference.}
Many studies investigate how scaling in inference affects LLM  performance. AlphaCode~\citep{li2022competition,leblond2023alphacode2} scales up the sample number and finds the solve rates scale log-linearly with more samples. ~\citet{brown2024large} improves LLMs' performance on math problems by repetitively sampling candidate solutions with a high temperature. ~\citet{wu2024empirical} and ~\citet{snell2024ScalingLLM} study the scaling behaviors of various inference time algorithms, reward models, and the model sizes. In this paper, we investigate the allocation algorithm between various inference configurations for scaling up inference.

\noindent \textbf{Inference Compute Optimization.}
There are two typical ways to optimize inference compute. 
One is to search for a single optimal configuration. For example, ~\citet{wang2023CostEffectiveHyperparameter} proposes EcoOptiGen to optimize the inference hyperparameter under a limited compute budget, and ~\citet{wang2024LLMCan} finetunes LLMs to self-regularize its generation with the best hyperparameter set. 
The other is to find the optimal allocation between various inference configurations~\citep{graves2016adaptive,dehghani2018universal}.
\citet{damani2024LearningHowHard} allocates the compute budget based on the estimation of problem difficulty. Here, we optimize the compute allocation to different inference configurations.

\begin{figure}[t!]
      \centering
		\includegraphics[width=1\linewidth]{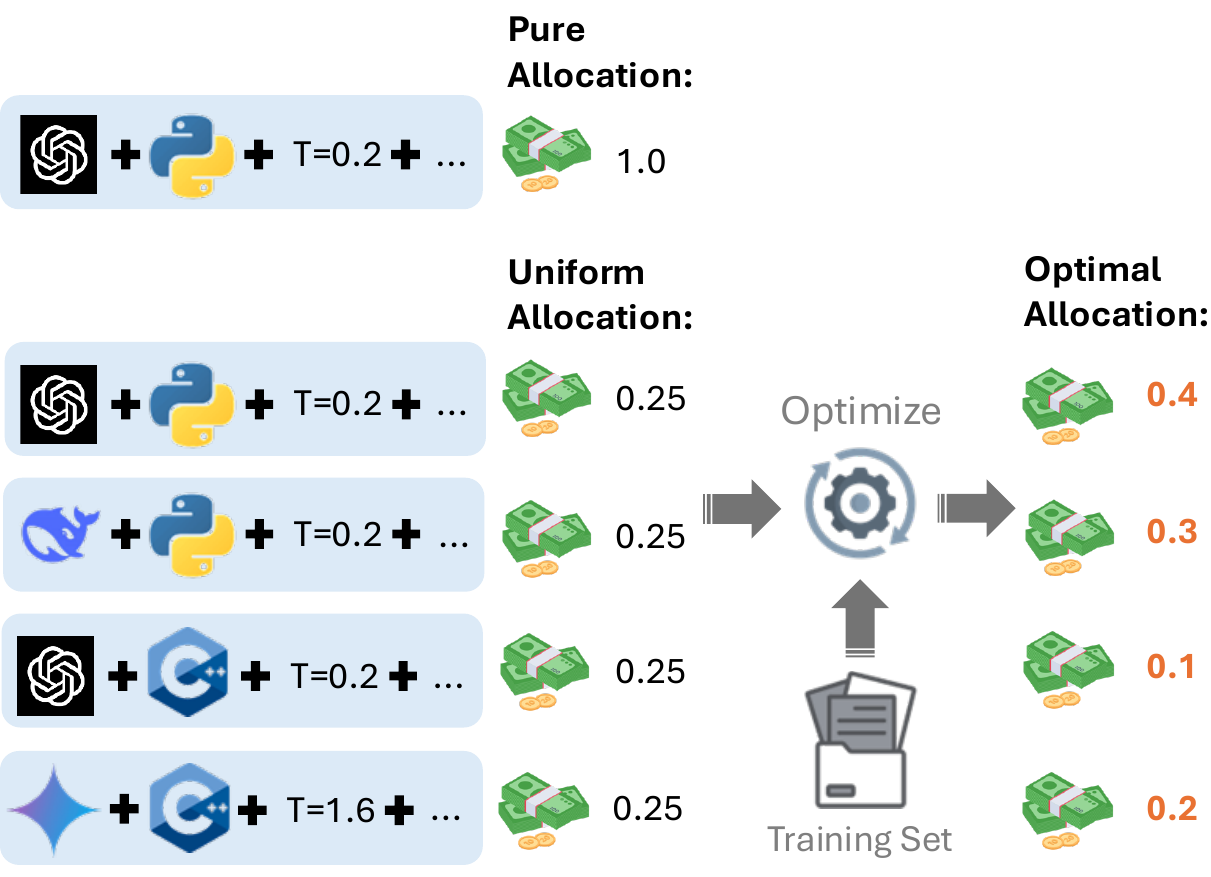}
 \caption{With 4 sampling configurations and a compute budget of 1, pure allocation spends all the budget on one configuration. Uniform allocation evenly distributes the budget across configurations. Optimized allocation learns where to spend more budget.}
\label{fig:strategies}
 \vspace{-16pt}
\end{figure}

\section{Method}

\subsection{Problem: Sample Compute Allocation}

\textbf{LLM Problem-Solving.}
We study how to improve LLM-based solvers for problems with binary correctness values.
Formally, a \textbf{\textit{problem}} is a pair $(x, v)$, where $x\in \mathcal{X}$ is the problem specifications, $v: \mathcal{X}\times \mathcal{Y} \rightarrow \{\texttt{True},\texttt{False}\}$ is the verifier that maps a solution of the problem to a truth value, and $\mathcal{Y}$ is the space of solutions.
An \textbf{\textit{LLM solver}} is given the problem specification and produces a distribution $\Pr_{\text{LM}}(y|x)$ over the solution space conditioned on the problem specification.

\noindent \textbf{Compute Budget and pass@$C$.}
We evaluate LLM-based problem solvers with the average solve rate of test set problems given a fixed \textbf{\textit{compute budget}} $C$, which can be defined as the maximum number of samples, the maximum number of tokens, the maximum FLOPs, etc.
For any definition of compute budget $C$, we generate as many samples as possible from $\Pr_{\text{LM}}(y|x)$ within the compute limit.
The solver can be evaluated using \textbf{\textit{pass@$C$}}, which is defined as the probability of at least one sample among the candidates being correct.
In our experiments, we use the number of samples as a metric for $C$.

\noindent \textbf{Sampling Configurations.}
Given a problem specification $x$, there are multiple inference hyperparameters to decide when solving it with an LLM -- the model to use, inference temperature, language of the output, prompt, etc.
For the $i$-th hyperparameter, we use $H_i$ to denote the set of feasible values and $h_i$ to denote an element in $H_i$ that is actually chosen.
Assuming the number of tunable hyperparameters is $d$, we call an $d$-tuple $\bm{h}=(h_1,h_2,\cdots,h_d) \in \mathcal{H}  = H_1\times H_2\times \cdots \times H_d$ a \textbf{\textit{sampling configuration}}.

\noindent \textbf{Sample Compute Allocation.}
Suppose there are $|\mathcal{H}| = m$ sampling configurations, and we want to allocate a compute budget $C$ across them.
We define an allocation as a mapping function $\pi: \mathcal{H} \rightarrow \mathbb{N}$ to represent the amount of compute assigned to each sampling configuration and it should satisfy $\sum_{\bm{h} \in \mathcal{H}} \pi(\bm{h}) = C$.
For convenience we use $\pi_i$ to denote $\pi(\bm{h}_i)$.
We categorize sample budget allocations into two types:
A \textbf{\textit{pure allocation}} spends all the compute on one sampling configuration, i.e., there exists exactly one $i$ for which $\pi_i>0$; while a \textbf{\textit{mixed allocation}} spends compute on more than one sampling configurations. 
Examples of different types of allocations can be found in Figure \ref{fig:strategies}.

\noindent \textbf{Learning Allocations.}
We want to find for a test set of problems $\mathcal{D}_{\text{test}}$ and a given per-problem compute budget $C$, a sample compute allocation $\pi$, such that the problem solve-rate pass@$C$ can be maximized given the same amount of compute.
We assume access to an i.i.d. training problem set $\mathcal{D}_{\text{train}}$ and a per-problem allocation-learning compute budget $C_0$ that can be used for trying out different sampling configurations.

\subsection{Why Mixed Allocation?}
\label{sec:example}

We show an example that demonstrates why it is sometimes necessary to use a mixed allocation of sample compute.
Consider two problems $x_1$ and $x_2$, and two configurations $\bm{h}_1$ and $\bm{h}_2$.
We use $\Pr(\text{pass}|\bm{h},x)$ to denote the probability of generating a correct solution to the problem $x$ under configuration $\bm{h}$.
Let's assume that $\Pr(\text{pass}|\bm{h_1},x_1)=10\%,\Pr(\text{pass}|\bm{h_2},x_1)=1\%,\Pr(\text{pass}|\bm{h_1},x_2)=1\%,\Pr(\text{pass}|\bm{h_2},x_2)=10\%$.
In other words, $\bm{h}_1$ is better at solving $x_1$ and $\bm{h}_2$ is better at solving $x_2$.

Consider the allocation of 10 samples per problem to these two configurations.
If we use a pure allocation, all 10 samples will be entirely given to either $\bm{h}_1$ or $\bm{h}_2$, the expected pass@10 would be
\begin{align*}
\frac{1 - (1-0.1) ^ {10} + 1 - (1-0.01)^{10}}{2}=37.3\%.
\end{align*}

On the other hand, if we use a mixed allocation and split 10 samples evenly between $\bm{h}_1$ and $\bm{h}_2$, the expected pass@10 would be
\begin{align*}
1 - (1-0.1)^5\times(1-0.01)^5=43.8\%,
\end{align*}
which is significantly higher than the pure allocation's result.

\subsection{Our Algorithm}

The proposed \method is described in Algorithm \ref{algo}.

\noindent \textbf{Estimating pass probability.}
For each sampling configuration $\bm{h}_i$ and each problem $x_j$ in the training set, we estimate a probability matrix $P$ with $|\mathcal{H}| \times |\mathcal{D}|$ elements. Each element $p_{ij}$ indicates the probability of a sample from $\bm{h}_i$ solving $x_j$.
We estimate $p_{ij}$ by sampling $C_0$ times from $\bm{h}_i$ and computing the frequency of correct solutions,
\begin{align}
p_{ij}\approx \frac{c_{ij}}{C_0},
\end{align}
where $c_{ij}$ is the number of correct samples generated for the problem $x_j$ with the configuration $\bm{h}_i$.
Note that to save compute, the estimation compute budget $C_0$ is much smaller than the actual compute budget $C$.

\noindent \textbf{Maximizing expected pass@$C$ on training set.}
Assuming that the test set is i.i.d. with the training set, we optimize $\pi$ to maximize pass@$C$ on training data.
For a single problem $x_j$ and a single configuration $\bm{h}_i \in \mathcal{H}$, its pass@$C_i$ is defined as it probability of being solved with compute $C_i$, which can be derived as
\begin{align}
\Pr(\text{$x_j$ solved with $C_i$ samples from $\bm{h}_i$}) \nonumber
\\
=1 - (1-p_{ij})^{C_i}.
\end{align}

By aggregating all problems and all configurations, we can obtain this optimization problem:
\begin{align}
\max_{\pi}\ & \E[\text{pass@}C]\nonumber\\
=&\frac{1}{|\mathcal{D}|}\sum_{j=1}^{|\mathcal{D}|}\left(1-\prod_{i=1}^{|\mathcal{H}|}(1-p_{ij})^{\pi_i} \right), \nonumber  \\
\text{s.t. } & 0\le \pi_i \le C \nonumber, \\
& \sum_{i=1}^{|\mathcal{H}|}\pi_i= C, \pi_i\in \mathbb{N}. \label{eq:opt}
\end{align}

Note that if we remove the integral constraint $\pi_i \in \mathbb{N}$ from Problem (\ref{eq:opt}), the relaxed problem is convex (Proof in Appendix \ref{sec:proof}) that can be solved optimally by hill climbing algorithm \citep{russell2016artificial}.
We start from a randomly picked distribution of compute $\pi^{(0)}$.
At each iteration $t$, we examine all the neighbors of the current distribution that differ slightly from $\pi^{(t)}$ and ``climb'' to the neighbor if it's better than the current distribution to obtain $\pi^{(t+1)}$.
The algorithm stops once there is no better neighbor.
Though the algorithm is not guaranteed to produce global optima for integral solutions, it works well empirically.

This algorithm can be extended to problems with real scores with some slight modification, as discussed in Appendix \ref{app:frac_score}.

\begin{algorithm}[ht]
\caption{\method: Algorithm for Optimizing Sample Compute Allocation}
\begin{algorithmic}[1]
\small
\State {\bfseries Input:} Inference budget $C$, estimation budget $C_0$, 
sampling configurations $\mathcal{H}$, 
training problem set $\mathcal{D}_{\text{train}}$, 
where $(x_i,v_i)\in D_{\text{train}}$ is a problem specification and its verifier.
\State {\bfseries Output:} Inference strategy $\pi$.

\State

\Function{OptimizeAllocation}{$C,C_0,\mathcal{H},\mathcal{D}_{\text{train}}$}
\State $P\leftarrow \textsc{EstimatePassProb}(C,C_0,\mathcal{H})$
\State Randomly initialize allocation $\pi=\{\pi_1, \dots, \pi_m\}$ such that $\sum_{i=1}^{m} \pi_i=C$.
\Repeat
    \State improved $\leftarrow$ \texttt{False}
    \State \textit{// Enumerating all neighboring strategies}
    \For{$i \leftarrow 1\text{ to }m$, $j \leftarrow 1\text{ to }m$ and $i\neq j$ and $\pi_i>0$}
    \State $q\leftarrow\pi$
    \State $q_i, q_j \leftarrow \pi_i - 1, \pi_j + 1$
  \If {\textsc{PredAcc}($q,P$)$>$\textsc{PredAcc($\pi,P$)}}
      \State $\pi \leftarrow q$ \textit{ // Climb to a better strategy.}
      \State improved $\leftarrow$ \texttt{True}.
  \EndIf
    \EndFor
\Until {\textbf{not} improved}
\EndFunction

\State

\State

\Function{EstimatePassProb}{$C,C_0,\mathcal{H}$}

\For {$\bm{h}_i \in \mathcal{H}$}
  \For {$(x_j, v_j) \in \mathcal{D}_{\text{train}}$}
     \State Generate $C_0$ samples $y_1,y_2,\cdots,y_{C_0}$, $y_i\sim \Pr_{\text{LM}}(y|x,\bm{h})$.
     \State $c\leftarrow \left| \{y_i:v_j(y_i)=\texttt{True}\}\right |$.
     \State $p_{ij}\leftarrow c/C_0$.
  \EndFor
\EndFor

\EndFunction

\State

\Function{PredAcc}{$\pi,P$}
    \State \Return $\frac{1}{|\mathcal{D}|}\sum_{j=1}^{|\mathcal{D}|}\left(1-\prod_{i=1}^{|\mathcal{H}|}(1-p_{ij})^{\pi(\bm{h}_i)}\right)$
\EndFunction
\end{algorithmic}\label{algo}
\end{algorithm}

\textbf{Discussion on the definition of \textit{compute budget}.}
In our experiments, we use the number of samples as a metric for compute, which is reasonable when the compute/price to generate each sample is similar for different configurations.
However, there are other cases when the model sizes differ greatly or when the FLOPs budget for each sampling configuration varies significantly.
For those cases, \textit{sample budget} is not a good proxy for \textit{compute budget}, and Problem (\ref{eq:opt}) will need to be modified to accommodate different definitions of compute budget, possibly by adding a coefficient to weight the sample budget for each problem.
Nonetheless, such modification does not change the convexity of the relaxed problem.
Thus our algorithm can still apply.

\section{Evaluation on Single-Turn Tasks}

We first evaluate \method on single-turn tasks, where a solution to a problem can be generated with a single LLM call.
For these tasks, the only way to scale up LLM inference is to sample more solutions.

\subsection{Baselines}

We consider three baselines in our experiments -- default pure allocation, optimal pure allocation, and uniform mixed allocation:

\textbf{Default pure allocation} is the one used to produce the leaderboard results on the benchmarks.
These benchmarks usually run LLM inference once for each problem instance, with a very basic prompt and a low temperature for reproducibility and fair comparison.
Therefore, the default pure allocation is not optimized for scaling up.

\textbf{Optimal pure allocation} is the one that has the highest pass@$C_0$ on the training set given a compute budget of $C_0$.
Since the actual $C$ is larger than $C_0$ in most cases, we use pass@$C_0$ to select the optimal configuration.
Comparison between the optimal pure allocation and the default pure allocation indicates whether it is necessary to search for a good set of inference hyperparameters, which is what existing hyperparameter optimization techniques do.

\textbf{Uniform mixed allocation} naively distributes the compute budget evenly to every sampling configuration $\bm{h}_i$ in $\bm{H}$.
We compare the uniform mixed allocation with our learned mixed allocation to examine whether it is necessary to optimize sample compute allocations for them to perform well.

\subsection{Tasks and Benchmarks}

We evaluate \method on five tasks from two benchmarks -- LiveCodeBench \citep{jain2024livecodebench} and LiveBench \citep{white2024livebench}.
These two benchmarks are built with periodically released examinations and competitions so that the possibility of contamination can be minimized.
To further avoid contamination, we choose the earlier released problems as training set $\mathcal{D}_{\text{train}}$.

\textbf{LiveCodeBench \citep{jain2024livecodebench}} collects LeetCode-style problems from weekly held online programming competitions such as LeetCode, Codeforces, and AtCoder.
Each problem comes with a natural language specification that includes problem description, sample test cases, and input range.
When given a problem, an LLM is supposed to create a program that can pass both the sample tests visible to the model and the hidden tests that are usually more comprehensive.
Within each category among the three (\textit{LeetCode, AtCoder, Codeforces}) in LiveCodeBench, we sort the problems according to the time they were released and use the first 3/5 of the problems as training data.
This split gives us 305 problems for training and 205 problems for evaluation.
Most training problems are released before 2024, while the evaluation problems are after 2024.

\textbf{LiveBench \citep{white2024livebench}} shares the same spirit as LiveCodeBench but extends the scope of problems.
There are 6 problem categories in LiveBench: \textit{data analysis, language, reasoning, math, instruction following, and coding}. The responses to the instruction-following problems are not easy to verify, and their coding problems are mostly taken from LiveCodeBench. The remaining four categories are used for evaluation. The problems have two release dates (\textit{2024-06-24 and 2024-07-26}), but most categories have only one release. Therefore, we mixed these two versions and randomly split each category by 3:7, resulting in 201 training and 471 testing problems.
Compared to LiveCodeBench, optimizing sample compute allocation is harder on LiveBench, because there are problems from different domains.

\begin{table}[hbpt]\footnotesize
  \centering
    \begin{tabular}{lcccc}
    \toprule
       &    & Model & Temp. & Lang. \\
    \midrule
    \multirow{5}[2]{*}{LCB} & Default & GPT-4o & 0.2 & Python \\
       & Optimal & GPT-4o & 1 & Python \\
       \cmidrule(lr){2-5}
       & \multirow{3}[1]{*}{Mixed} & GPT-4o & \multirow{3}[1]{*}{0-1.6} & \multirow{3}[1]{*}{Python, C++} \\
       & & Gemini & &\\
       & & DeepSeek & &\\
    \midrule
    \multirow{5}[2]{*}{LB} & Default & Llama3   & 0   &  N/A \\
       & Optimal & Llama3  & 0.8  & N/A  \\
       \cmidrule(lr){2-5}
       & \multirow{3}[1]{*}{Mixed} & Qwen2 & \multirow{3}[1]{*}{0-1.8} & \multirow{3}[1]{*}{N/A} \\
       & & Llama3 & &\\
       & & DeepSeek & &\\
    \bottomrule
    \end{tabular}%
  \caption{Implementation details for the two benchmarks, including the hyperparameters we consider, the default inference strategy on the leaderboard, and the budget for strategy learning and inference. Temp. and Lang. are the temperature and the programming language used when sampling. }
  \label{tab:setting}%
\end{table}%

\begin{table*}[ht]
\small
\centering
\begin{tabular}{llr*{8}{r}}
\toprule
    &   & \multicolumn{9}{c}{pass@$C$ with varying $C$ $\uparrow$}    \\ 
    \midrule
    &   &  $2^1$ & $2^2$ & $2^3$ & $2^4$ &$2^5$ & $2^6$&$2^7$ &$2^8$& $2^9$\\ 
    \midrule
    \multirow{4}{*}{LiveCodeBench}  
    & Default pure    & 47.1 & 49.4 & 51.7 & 53.9 & 55.7 & 57.1 & 58.4 & 59.5 & 60.8   \\
    & Optimal pure    & +1.8 & +4.1 & +5.6 & +6.6 & +7.2 & +7.5 & +7.6 & +8.4 & +9.4 \\
    & Uniform mixed & \underline{+4.3} & \underline{+15.0} & \underline{+21.6} & \underline{+24.4} & \underline{+25.8} & \underline{+26.7} & \underline{+27.2} & \textbf{+27.6} & \textbf{+28.1}   \\
    & Learned mixed (ours) & \textbf{+17.4} & \textbf{+22.6} & \textbf{+23.7} & \textbf{+25.2} & \textbf{+26.3} & \textbf{+27.3} & \textbf{+27.7} & \underline{+27.2} & \underline{+26.9}  \\ 
    \midrule
    \multirow{4}{*}{LiveBench}  
    & Default pure    & 46.0 & 51.3 & 55.9 & 59.7 & 62.6 & 64.9 & 66.7 & 68.3 & -  \\
    & Optimal pure    & \underline{+2.4} & \underline{+6.0} & \underline{+9.3} & \underline{+10.0} & \underline{+11.2} & \underline{+12.5} & \underline{+13.8} & \underline{+14.7} & -  \\
    & Uniform mixed   & -11.7 & -4.4 & +3.1 & +8.0 & +10.5 & +11.6 & +12.5 & +12.8 & -  \\
    & Learned mixed (ours) & \textbf{+5.9} & \textbf{+8.8} & \textbf{+11.1} & \textbf{+12.3} & \textbf{+13.0} & \textbf{+13.8} & \textbf{+14.8} & \textbf{+14.7} & -  \\ 
    \bottomrule
\end{tabular}
\caption{Accuracy at different compute budgets ($C$) using different sample compute allocations. We report the difference in accuracy for non-default allocations. LCB stands for code generation tasks in LiveCodeBench. LB stands for subtask categories in LiveBench.
The best result for each setting is in bold.}
\label{tab:results}
\end{table*}

We list the sampling configurations in Table \ref{tab:setting}. The temperatures we consider are multiples of 0.2. For all configurations, we set $C_0$ to 50.
Note that there is one more hyperparameter to optimize for LiveCodeBench -- the programming language.
For LiveBench, the open-weight models we choose all have about 70B parameters.
We also provide the values of hyperparameters for the pure allocation baselines and the budgets for learning and testing.

\subsection{Key Observations}

We present the pass rates for different sample compute allocations in Table \ref{tab:results}. For allocations other than the default pure, we show the difference between their pass rate and that of default pure. Several key observations can be made from these results:

\textbf{Pure allocation is not enough, mixed allocation is necessary.}
By finding the optimal pure allocation on the training set, we get much better accuracy than the default allocation, highlighting the importance of a suitable inference configuration. However, pure allocation is not enough on LiveCodeBench. We find that code problems require a more diverse solution set, making the pure configuration not enough.
On LiveCodeBench, just by allocating sample compute evenly across inference settings, we achieve a pass@8 of 73.3\%, which is better than the pass rate of optimal pure allocation with 512 samples.

\textbf{Uniform mixed allocation is not enough, \method's optimized allocation is necessary.}
On LiveBench, uniform mixed allocation does not outperform optimal pure, suggesting that we can't always opt for uniform mixed.
However, \method's optimized allocation does.
With 8 samples, \method's pass rate comes to 67\%, which is higher than the pass rate of the default pure allocation with 128 samples.
In fact, \method outperforms all others in all but two cases.

\textbf{\method scales well with larger compute budgets.}
Although the sample compute budget $(C_0)$ used for estimating the solve rate matrix was merely 50, \method can still produce strong sample compute allocations even when the test time compute budget $C$ is as large as 1024 per problem.

\begin{figure}[th!]
      \centering
		\includegraphics[width=0.9\linewidth]{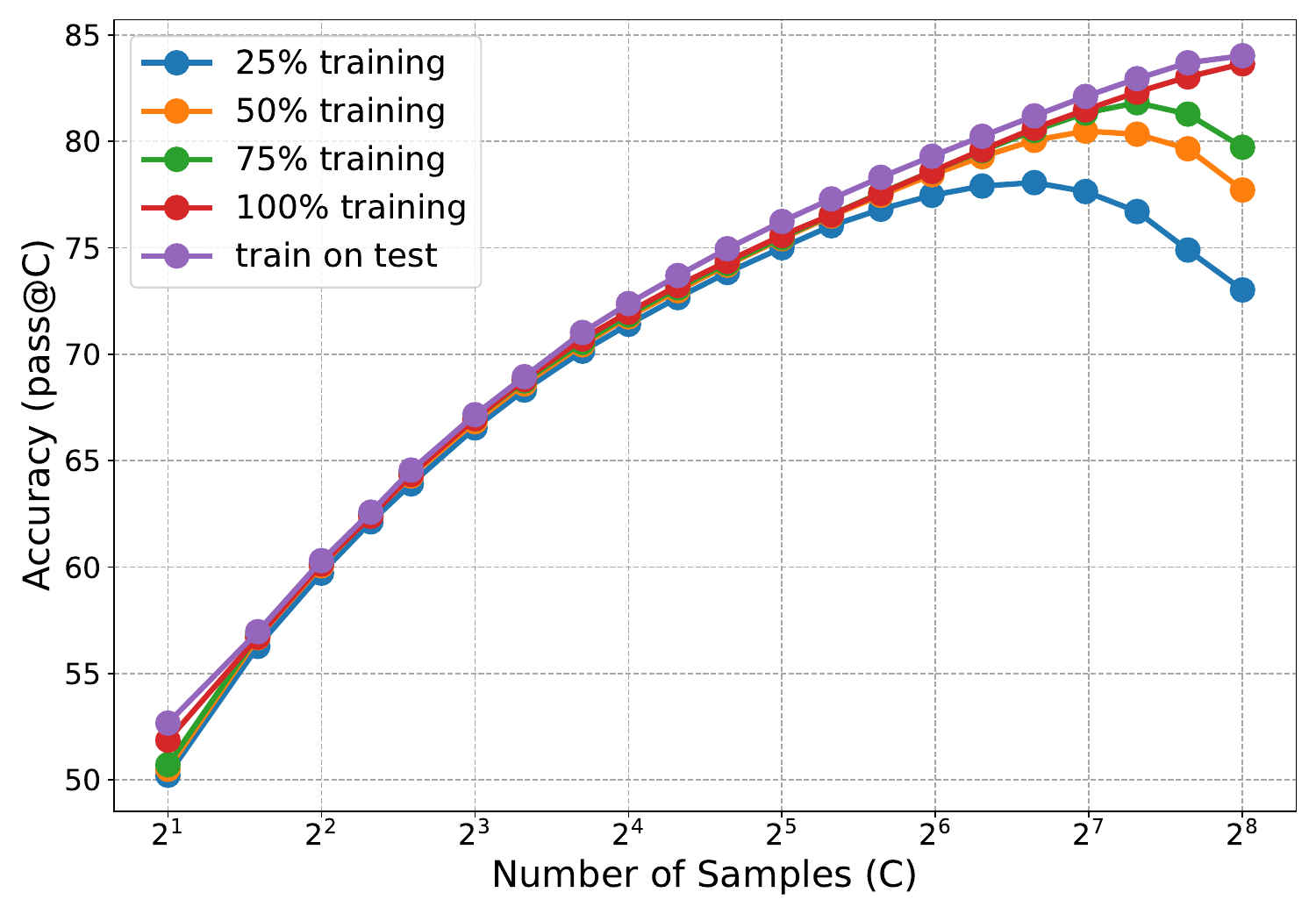}
 \caption{\method's pass rates on LiveBench when trained with different proportions of the training data.}
\label{fig:train_abl}
\end{figure}

\subsection{Ablation Studies}
\label{sec:abl}

We conduct ablation experiments on LiveBench to answer the following questions.

\textbf{How many problems do we need in training to learn a good mixed allocation?}
We run \method on different proportions of the original training set and plot the results in Figure \ref{fig:train_abl}.
Since there are multiple ways of subsampling the training set, we run it multiple times and compute the average.
For reference, we also plot the results when we train on the test set, which should be the upperbound of \method.
We observe that with the full training data, the allocation learned is very close to the upperbound.
At smaller sample compute $C$, the allocation trained from less data is not much worse than the allocation from full training data.
However, as $C$ gets to over $2^6$, the allocation gets much worse with less training data.
We hypothesize that this is because when $C$ is large, only the hard problems contribute to the difference in accuracy, which are prone to overfitting when training data is small.

\begin{figure}[th!]
      \centering
		\includegraphics[width=0.9\linewidth]{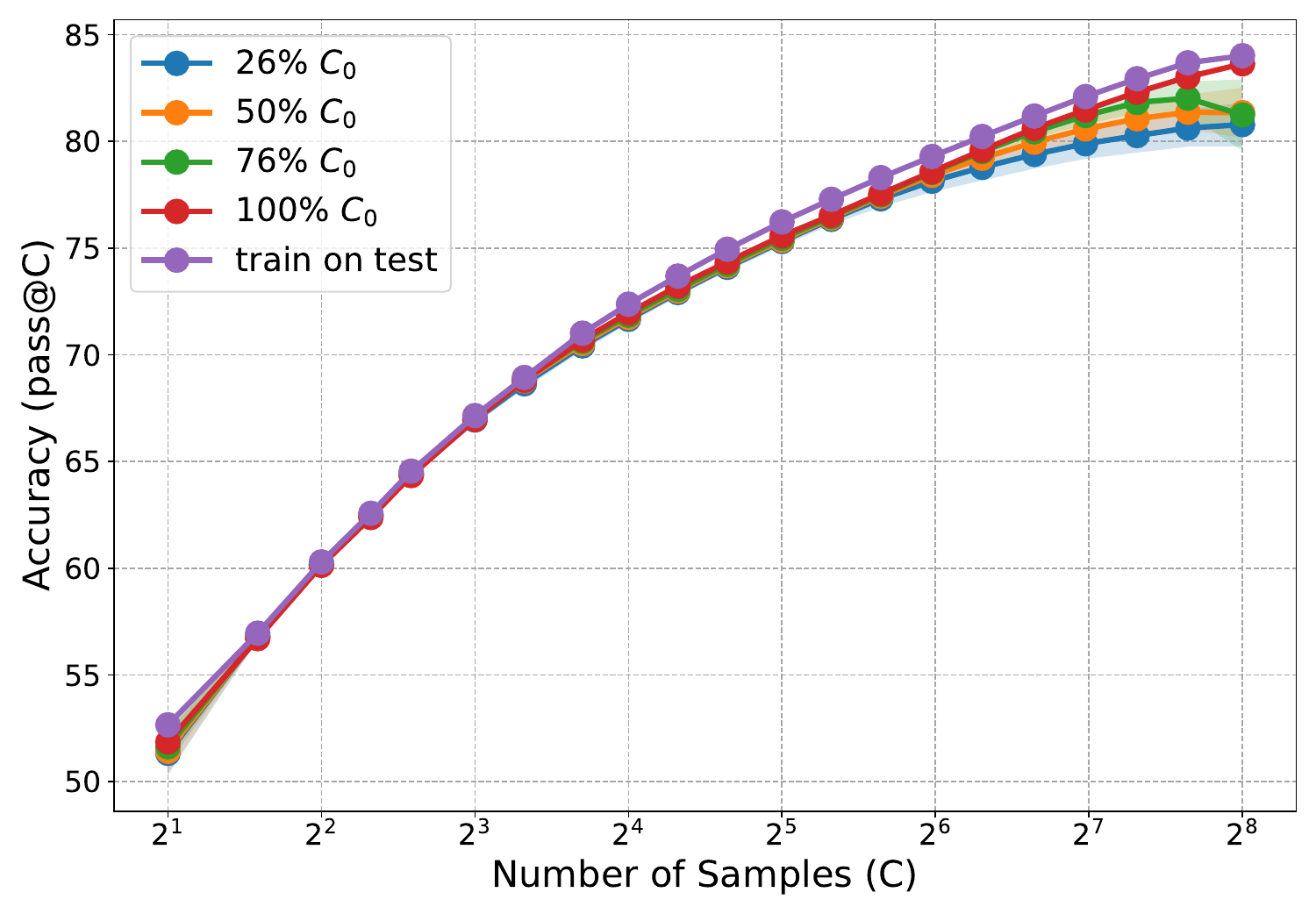}
 \caption{\method with different sample compute for estimating pass rates. $C_0=50$.}
\label{fig:c0_abl}
\end{figure}

\begin{figure}[th!]
      \centering
		\includegraphics[width=0.9\linewidth]{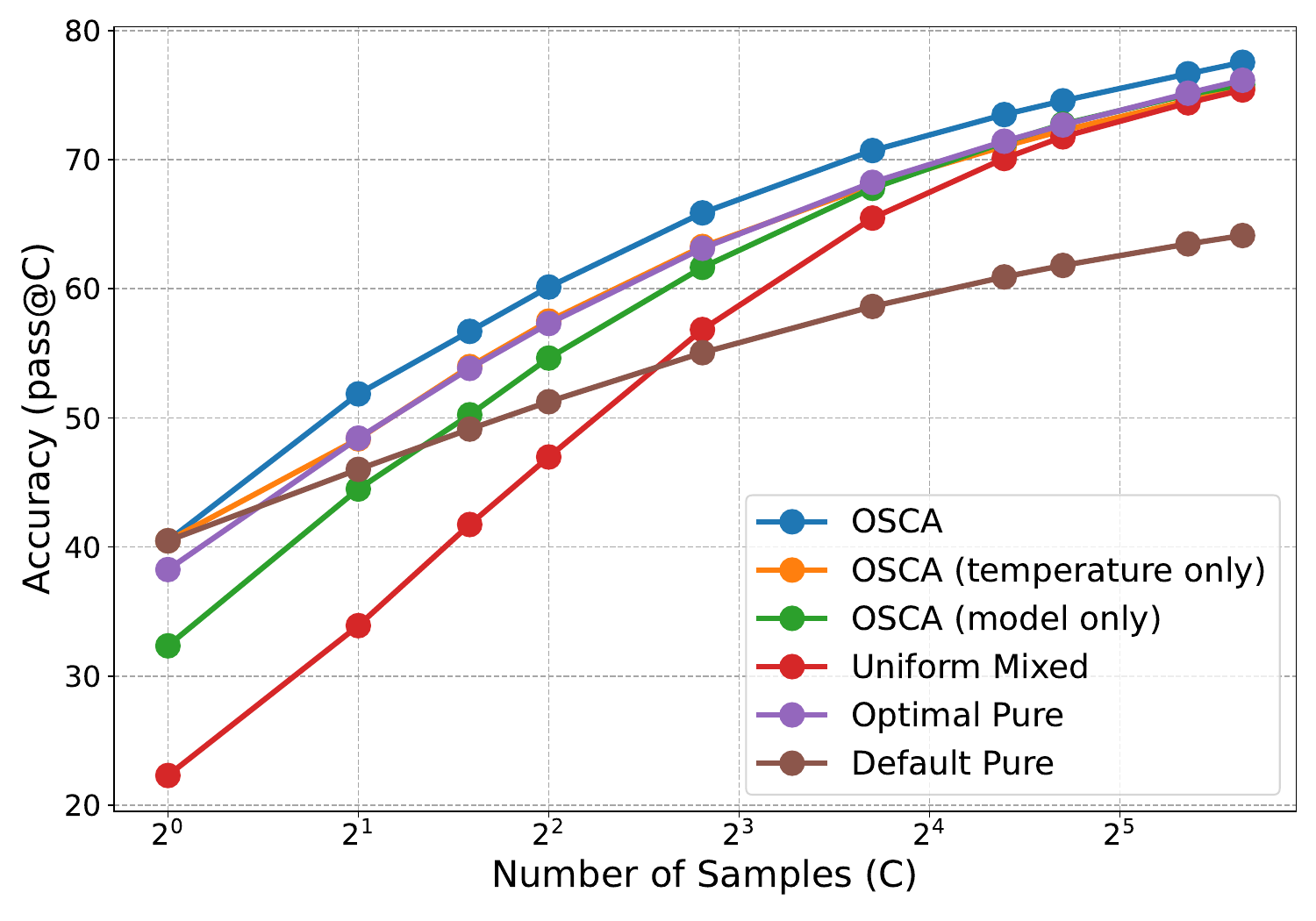}
 \caption{\method's pass rates on LiveBench when it is banned from allocating compute to multiple temperatures or multiple models.}
\label{fig:hyp_abl}
\end{figure}

\textbf{How large does $C_0$ need to be in order to estimate pass rates on training data?}
$C_0$, the sample compute budget for estimating pass rates on training data, can be much smaller than $C$, which might affect \method's performance.
We study how estimating with different values of $C_0$ might affect the learning process.
As Figure~\ref{fig:c0_abl} shows, even with 26\% of the original $C_0$, \method can still get an accuracy very close to its upperbound, indicating that it is not really sensitive to $C_0$.

\textbf{Which hyperparameter in sampling configurations needs mixed allocation?}
To study whether mixed allocation is needed for the two hyperparameters -- temperature or model, we consider two limited versions of \method by banning it from using more than one temperature or more than one model.
As shown in Figure~\ref{fig:hyp_abl}, when \method is not allowed to use multiple models, its accuracy degrades to that of optimal pure.
When \method is not allowed to use multiple temperatures, its accuracy degrades to be worse than optimal pure.
These results suggest that in order for \method to perform well, we need more diverse sampling configurations.

\begin{table}[ht]
\small
\centering
\begin{tabular}{llr*{3}{r}}
\toprule
    &   & \multicolumn{3}{c}{pass@$C$ with varying $C$ $\uparrow$}    \\ 
    \midrule
    &   & $2^3$ & $2^5$ & $2^7$ \\ 
    \midrule
    \multirow{4}{*}{T1}  
    & Default pure   & 53.7  & 60.3  & 65.2   \\
    & Optimal pure      & \underline{+7.0}  & \underline{+9.1}  & \underline{+12.4}  \\
    & Uniform mixed    & +2.5 & +8.1& +9.7  \\
    & Learned mixed  & \textbf{+10.3} & \textbf{+11.4}  & \textbf{+14.0}  \\ 
    \midrule
    \multirow{4}{*}{T2} 
    & Default pure  & 70.1  & 82.5 & 87.5 &  \\
    & Optimal pure    & \underline{+6.0} & +5.4 & +6.2   \\
    & Uniform mixed  & +1.4 & \textbf{+10.6} & \textbf{+9.7} \\ 
    & Learned mixed & \textbf{+12.1} & \underline{+10.1} & \underline{+8.8}  \\
    \midrule
    \multirow{4}{*}{T3}  
    & Default pure    & 47.4 & 54.5 & 59.9 &  \\
    & Optimal pure     & \underline{+8.9}  & \underline{+15.1}  & \textbf{+17.8}   \\ 
    & Uniform mixed   & +2.5 & +10.0 & +12.6  \\
    & Learned mixed   & \textbf{+14.1} & \underline{+16.1} & \underline{+16.2}  \\ 
    \midrule
    \multirow{4}{*}{T4}  
    & Default pure    & 52.5 & 53.2 & 53.9 \\
    & Optimal pure    & \textbf{+9.8} & \underline{+16.2} & +20.0  \\
    & Uniform mixed  & +6.4 & +15.4 &  \underline{+20.6}  \\
    & Learned mixed & \underline{+9.3} & \textbf{+16.3} & \textbf{+21.9} \\ 
    \bottomrule
\end{tabular}
\caption{Accuracy of different allocations on subtasks of LiveBench -- math (T1), reasoning (T2), language (T3), data analysis (T4).}
\label{tab:abl_lb}
\end{table}

\textbf{How general is \method's learned allocation?}
In Table \ref{tab:results}, we report the overall performance on LiveBench, because \method's training data is a combination of 4 subtasks from LiveBench -- math, reasoning, language and data analysis.
To evaluate the generality of the learned allocation, we examine its performance on the 4 tasks separately.
As Table~\ref{tab:abl_lb} shows, \method is the best in 7/12 cases and the second best in the remaining 5.
This indicates that the \method's learned allocation is domain-specific to some extent.
Example problems of these 4 categories can be found in Appendix~\ref{app:lb_samples}.

\textbf{Can we learn the best configuration at instance level?}
Ideally, there is no need to use mixed allocation of sampling compute because there is only one optimal configuration for a specific problem.
Therefore, we investigate whether it is possible to learn instance-level optimal configuration.
We conduct this experiment on LiveCodeBench, as it has a larger training set and more homogeneous problems and is thus easier to learn.
We propose a k-nearest neighbor algorithm under the assumption that similar problems need similar sampling configurations.
For each test problem $x$, we retrieve the $k$ problems in the training set that are most similar to $x$, and allocate compute according to the distribution of optimal configuration over these problems.
We use OpenAI's \texttt{text-embedding-3-large} to create semantic embeddings and compute the similarity between problems.
As shown in Table~\ref{tab:knn}, \method outperforms the problem-specific kNN allocation of compute, no matter what value $k$ is.
This suggests that it is not straightforward to learn problem-specific sampling configurations, justifying the need for a domain-specific allocation.

\begin{table}[ht]
\small
\centering
\begin{tabular}{lr*{4}{r}}
\toprule
   & \multicolumn{4}{c}{pass@$C$ with varying $C$ $\uparrow$}    \\ 
    \midrule
     &  $2^1$ & $2^3$ & $2^5$ & $2^7$ \\ 
    \midrule 
     $k=4$   & 65.7  & 73.7  & 80.0 & 83.9   \\
     $k=8$   & 66.9  & 74.5  & 79.9 & 82.9  \\
     $k=16$      & 68.8  & 76.6  & 83.2 & 86.1 \\
     $k=32$    & 68.2 & 77.3 & 83.0 & 86.6  \\
     $k=64$  & 68.0 & 77.2 & 83.4 & 86.2 \\
     \method (ours) & \textbf{72.0} & \textbf{79.1} & \textbf{84.4} & \textbf{86.7} \\
    
    \bottomrule
\end{tabular}
\caption{Accuracy of problem-specific sample allocation computed using k-nearest neighbor under different values of $k$ is consistently lower than \method, especially when the compute budget is small.}
\label{tab:knn}
\end{table}

\section{Evaluation on Agentic Tasks}

Optimizing sample compute allocation is not just useful for its own purpose, it is also useful for more complicated LLM inference algorithms, such as tree search and agentic workflows, because sampling is a basic operation in these.
To demonstrate such usefulness, we apply \method to an agentic workflow for SWE-Bench, which needs 2 stages and 7 steps involving numerous LLM calls to resolve software engineering issues in repository-level code.

\subsection{Benchmark and Workflow}

\noindent \textbf{SWE-Bench \citep{jimenezswe}} collects real-world software engineering issues from open-source GitHub repositories such as django and matplotlib.
Each issue is paired with human-written unit tests.
To resolve an issue, one needs to understand the issue description, examine the codebase (often hundreds of thousands of lines long), locate where to make changes and make necessary modifications by generating a patch.
Decent solutions on this benchmark often takes an agentic approach by giving an LLM multiple tools to use and multiple actions to take and guiding it through a multi-stage workflow.
We consider a subset of SWE-Bench curated by the authors called SWE-Bench Lite, which contains 300 issues.

\noindent \textbf{Agentless \citep{xia2024agentless}} is one of the best-performing open-source solutions to SWE-Bench.
An overview of how Agentless works can be found in Appendix \ref{app:agentless workflow} Figure \ref{fig:agentless}.
It needs two stages -- bug localization and bug repairing -- to resolve a software engineering issue.
There are 7 steps in total and 4 of them needs LLM sampling.
We apply \method to one crucial step in the workflow -- generating patches.
Since SWE-Bench is really expensive to evaluate, we set both $C_0$ and $C$ to be a smaller value 16.

\noindent \textbf{Implementation Details.} We use the same baselines from single-turn tasks -- default pure allocation, optimal pure allocation, and uniform mixed allocation.
We randomly sample 150 from the 300 issues as \method's training set, and use the remaining ones for testing.
We consider three dimensions in sampling configuration -- model choice, temperature, and prompting.
The model choices we consider are \textit{gpt-4o-2024-05-13}, \textit{deepseek-coder-2.5}, and \textit{qwen-2.5-70B}.
The temperatures we consider are 0.4, 0.8, 1.2, and 1.6.
The prompts are the intermediate results generated from the earlier stages of the workflow.
We consider the two sets of intermediate results provided by the authors of Agentless\footnote{
The prompt sets can downloaded at \url{https://github.com/OpenAutoCoder/Agentless/releases/tag/v0.1.0}.
Prompt set 1 is in \url{results/location\_merged/loc\_merged\_0-1\_outputs.jsonl}.
Prompt set 2 is in \url{results/location\_merged/loc\_merged\_2-3\_outputs.jsonl}.
}.
They contain the relevant contexts in code files that the system finds relevant to the specific issue descriptions.

Both the default and optimal settings are gpt-4o, temperature 0.8 and prompt set 1.

\subsection{Results}

As demonstrated in Figure \ref{fig:fig1}, \method can also improve the performance of Agentless by improving one of the steps in its agentic workflow.
Compared to both the pure allocation (Temperature is set to 1 in both default and optimal pure allocation) and uniform mixed allocation, \method's learned allocation can get a similar accuracy with fewer samples.
The gap between \method and the optimal pure allocation enlarges as sample compute gets larger,
while uniform mixed allocation approaches and outperforms optimal pure with larger sample sizes.

These findings further demonstrate the necessity of optimizing sample budget allocation, as it can be used in more complicated workflows to make them more efficient.

\section{Conclusion}

In conclusion, this paper presents \method, an algorithm designed to optimize sample compute allocation for large language models (LLMs) during inference. Through various experiments on both single-turn and agentic tasks, the study demonstrates that \method significantly improves accuracy with reduced compute resources compared to traditional methods, such as pure or uniform mixed allocations. By leveraging a mixed allocation allocation, \method balances different inference configurations, proving particularly effective in code generation and reasoning tasks, as seen in benchmarks like LiveCodeBench and LiveBench.
This work demonstrates the importance of adapting sampling allocation to the specific characteristics of the problem at hand. Furthermore, \method's application to more complex workflows, such as multi-step agentic tasks, shows potential for broader utility in improving the efficiency of LLMs in various real-world applications. However, future work could explore optimizing additional hyperparameters and testing the scalability of the method with larger compute budgets.

\section*{Limitation}
Although \method demonstrates an effective way to allocate sample compute there are still several limitations. First, this paper mainly focuses on four representative inference hyperparameters: model types, temperatures, response languages, and prompts. In addition to these aspects, there are other hyperparameters such as top k, top p, repetition penalty, etc. Combining these hyperparameters can make the sample configuration set more diverse. Besides, due to the computation limitation, we limited our inference compute budget to 512. It would be interesting to see how further scaling up will affect the performance. 

\section*{Ethics Statement}

We acknowledge that there might be some ethical considerations in enhancing LLMs' capability such as the \method presented in this paper. However, we believe that none must be specifically highlighted here.

\bibliography{custom}

\newpage

\appendix

\section{Appendix}
\label{sec:appendix}

\subsection{Proof of Convexity}
\label{sec:proof}

We aim to minimize the following objective function:

\[
\min_{\pi} O(\pi) = \sum_{j=1}^m \prod_{i=1}^n (1 - p_{ij})^{\pi_i}
\]

subject to:

\begin{itemize}
    \item \( 0 \leq \pi_i \leq C \), with \( \pi_i \in \mathbb{N} \)
    \item \( \sum_{i=1}^n \pi_i = C \)
\end{itemize}

To simplify the analysis, we take the negative logarithm of the objective function, which preserves convexity properties:

\[
O(\pi) = \sum_{j=1}^m \exp\left( \sum_{i=1}^n \pi_i \ln(1 - p_{ij}) \right)
\]

Define:

\[
g_j(\pi) = \sum_{i=1}^n \pi_i \ln(1 - p_{ij})
\]

Since \(g_j(\pi\) is a linear combination of \(\pi_i\), it is an affine function in \(\pi\). As affine functions are both convex and concave, \(g_j(\mathbf{\pi})\) is convex.

The exponential function, \(\exp(z)\), is convex. Since the composition of a convex function with an affine function remains convex, it follows that:
\begin{equation}
h_j(\pi) = \exp(g_j(\pi)) \nonumber
\end{equation}

is convex in \(\pi\).

Thus, the overall objective function:
\begin{align*}
f(\pi)& = \sum_{j=1}^m h_j(\pi)\\
&= \sum_{j=1}^m \exp\left( \sum_{i=1}^n \pi_i \ln(1 - p_{ij}) \right) \nonumber
\end{align*}
is a sum of convex functions, implying that \(f(\pi)\) is convex.

\qed

\subsection{Example Problems from LiveCodeBench}

\begin{examplebox}[Codeforces: Short Sort]

\footnotesize
\textbf{Problem Description:}
There are three cards with letters \texttt{a}, \texttt{b}, and \texttt{c} placed in a row in some order. You can do the following operation at most once:
\begin{itemize}
    \item Pick two cards and swap them.
\end{itemize}
Is it possible that the row becomes \texttt{abc} after the operation? Output \texttt{YES} if it is possible, and \texttt{NO} otherwise.

\textbf{Input:}

The first line contains a single integer $t$ ($1 \leq t \leq 6$) — the number of test cases.

The only line of each test case contains a string consisting of the characters \texttt{a}, \texttt{b}, and \texttt{c} exactly once, representing the cards.

\textbf{Output:}

For each test case, output \texttt{YES} if you can make the row \texttt{abc} with at most one operation, or \texttt{NO} otherwise.

\textbf{Example:}
\begin{verbatim}
Input:
6
abc
acb
bac
bca
cab
cba
Output:
YES
YES
YES
NO
NO
YES
\end{verbatim}

\textbf{Python Solution:}
\begin{verbatim}
def is_possible_to_sort_to_abc(s):
    if s == "abc":
        return "YES"
    if s == "acb" or s == "bac" or s == "cba":
        return "YES"
    return "NO"
t = int(input())
for _ in range(t):
    s = input().strip()
    print(is_possible_to_sort_to_abc(s))
\end{verbatim}
    
\textbf{C++ Solution:}
\begin{verbatim}
#include <iostream>
#include <string>
using namespace std;
int main() {
    int t;
    cin >> t;
    while (t--) {
        string s;
        cin >> s;
        if (s == "abc" || s == "acb"
        || s == "bac" || s == "cba") {
            cout << "YES" << endl;
        } else {
            cout << "NO" << endl;
        }
    }
    return 0;
}
\end{verbatim}

\end{examplebox}

\begin{examplebox}[LeetCode: Number of Senior Citizens]
\textbf{Problem Description:}
You are given a 0-indexed array of strings \texttt{details}. Each element provides information about a passenger compressed into a string of length 15. The format of the string is:
\begin{itemize}
    \item The first 10 characters consist of the phone number of the passengers.
    \item The next character denotes the gender of the person.
    \item The following two characters indicate the age of the person.
    \item The last two characters determine the seat allotted to that person.
\end{itemize}
Return the number of passengers who are strictly more than 60 years old.

\textbf{Example:}
\begin{verbatim}
Input: details = ["7868190130M7522", 
                  "5303914400F9211", 
                  "9273338290F4010"]
Output: 2

Explanation: The passengers at indices 0, 1,
and 2 have ages 75, 92, and 40. Thus, there
are 2 people who are over 60 years old.
\end{verbatim}

\textbf{Python Solution:}
\begin{verbatim}
class Solution(object):
    def countSeniors(self, details):
        count = 0
        for detail in details:
            age = int(detail[11:13])
            if age > 60:
                count += 1
        return count
\end{verbatim}

\textbf{C++ Solution:}
\begin{verbatim}
class Solution {
public:
    int countSeniors(vector<string>& details) {
        int count = 0;
        for (const string& detail : details) {
            int age = stoi(detail.substr(11, 2));
            if (age > 60) {
                count++;
            }
        }
        return count;
    }
};
\end{verbatim}

\end{examplebox}

\begin{examplebox}[AtCoder: Wrong Answer]
\textbf{Problem Description:}
You are given two integers $A$ and $B$, each between 0 and 9, inclusive. Print any integer between 0 and 9, inclusive, that is not equal to $A + B$.

\textbf{Input:}

The input consists of two integers $A$ and $B$.

\textbf{Output:}

Print any integer between 0 and 9, inclusive, that is not equal to $A + B$.

\textbf{Example:}
\begin{verbatim}
Input:
2 5

Output:
2

Input:
0 0

Output:
9

Input:
7 1

Output:
4
\end{verbatim}

\textbf{Python Solution:}
\begin{verbatim}
A, B = map(int, input().split())
S = A + B
for i in range(10):
    if i != S:
        print(i)
        break
\end{verbatim}

\textbf{C++ Solution:}
\begin{verbatim}
#include <iostream>
using namespace std;

int main() {
    int A, B;
    cin >> A >> B;
    int sum = A + B;
    for (int i = 0; i <= 9; ++i) {
        if (i != sum) {
            cout << i << endl;
            break;
        }
    }
    return 0;
}
\end{verbatim}

\end{examplebox}

\subsection{Example Problems from LiveBench}
\label{app:lb_samples}

\begin{examplebox}[Math Problem]
Real numbers $x$ and $y$ with $x, y > 1$ satisfy $\log_x (y^x) = \log_y (x^4y) = 10$. What is the value of $xy$? Please think step by step, and display the answer at the very end of your response. The answer is an integer consisting of exactly 3 digits (including leading zeros), ranging from 000 to 999, inclusive. \\

Ground Truth: 025
\end{examplebox}

\begin{examplebox}[Reasoning Problem]
In this question, assume each person either always tells the truth or always lies. Tala is at the movie theater. The person at the restaurant says the person at the aquarium lies. Ayaan is at the aquarium. Ryan is at the botanical garden. The person at the park says the person at the art gallery lies. The person at the museum tells the truth. Zara is at the museum. Jake is at the art gallery. The person at the art gallery says the person at the theater lies. Beatriz is at the park. The person at the movie theater says the person at the train station lies. Nadia is at the campground. The person at the campground says the person at the art gallery tells the truth. The person at the theater lies. The person at the amusement park says the person at the aquarium tells the truth. Grace is at the restaurant. The person at the aquarium thinks their friend is lying. Nia is at the theater. Kehinde is at the train station. The person at the theater thinks their friend is lying. The person at the botanical garden says the person at the train station tells the truth. The person at the aquarium says the person at the campground tells the truth. The person at the aquarium saw a firetruck. The person at the train station says the person at the amusement park lies. Mateo is at the amusement park. Does the person at the train station tell the truth? Does the person at the amusement park tell the truth? Does the person at the aquarium tell the truth? Think step by step, and then put your answer in **bold** as a list of three words, yes or no (for example, **yes, no, yes**. If you don’t know, guess. \\

Ground Truth: no, yes, yes
\end{examplebox}

\begin{examplebox}[Language Problem]
You are given 8 words/phrases below. Find two groups of four items that share something in common. Here are a few examples of groups: bass, flounder, salmon, trout (all four are fish); ant, drill, island, opal (all four are two-word phrases that start with ’fire’); are, why, bee, queue (all four are homophones of letters); sea, sister, sin, wonder (all four are members of a septet). Categories will be more specific than e.g., ’5-letter-words’, ’names’, or ’verbs’. There is exactly one solution. Think step-by-step, and then give your answer in **bold** as a list of the 8 items separated by commas, ordered by group (for example, **bass, founder, salmon, trout, ant, drill, island, opal**). If you don’t know the answer, make your best guess. The items are: row, drift, curl, tide, current, press, fly, wave. \\

Ground Truth: current, drift, tide, wave, curl, fly, press, row
\end{examplebox}

Example questions from the Data Analysis category can be lengthy, so examples can be viewed \href{https://huggingface.co/datasets/livebench/data_analysis/viewer/default/test?row=0}{here}.

\subsection{Mathematical Formulas for Evaluating Results with Fractional Scores}
\label{app:frac_score}

\textbf{Probability Mass Function (PMF) for the Maximum Score When Sampling \( k \) Scores}

To calculate the probability that the maximum score \( X_{\text{max}} \) among \( k \) samples is exactly \( x \):

\[
P(X_{\text{max}} = x) = \frac{\binom{c_{\leq x}}{k} - \binom{c_{< x}}{k}}{\binom{m}{k}}
\]

Where:
\begin{itemize}
    \item \( m = \sum_{x} c_x \) is the total sample size.
    \item \( c_{\leq x} = \sum_{y \leq x} c_y \) is the cumulative count of scores less than or equal to \( x \).
    \item \( c_{< x} = \sum_{y < x} c_y \) is the cumulative count of scores strictly less than \( x \).
\end{itemize}

\textbf{Expected Maximum Score Across Multiple Settings with Known PMF}

The expected value \( E[X] \) of the maximum score across multiple settings is:

\[
E[X] = \sum_{x} x \cdot P(X = x)
\]

Where \( P(X = x) \) is computed as:

\[
P(X = x) = \prod_{j=1}^{s} P_j(X_j \leq x) - \prod_{j=1}^{s} P_j(X_j < x)
\]

Here, \( P_j(X_j \leq x) \) represents the probability that the maximum score in setting \( j \) is less than or equal to \( x \). The difference between the products isolates the probability that the maximum score is exactly \( x \).

\textbf{Expected Maximum Score with Excess Samples}

The expected value of the maximum score when sampling \( n \) times is estimated by:

\[
E[X] = \sum_{x} \left( x \cdot \left[ \left( c + f_x \right)^{n} - c^{n} \right] \right)
\]

Where:
\begin{itemize}
    \item \( n \) is the number of samples,
    \item \( f_x \) is the probability density for score \( x \),
    \item \( c \) is the cumulative probability up to score \( x \).
\end{itemize}

The term \( (c + f_x)^{n} - c^{n} \) reflects the probability of selecting exactly \( x \) as the maximum score from \( n \) samples.

\newpage
\begin{figure*}[th!]
      \centering
		\includegraphics[width=0.9\linewidth]{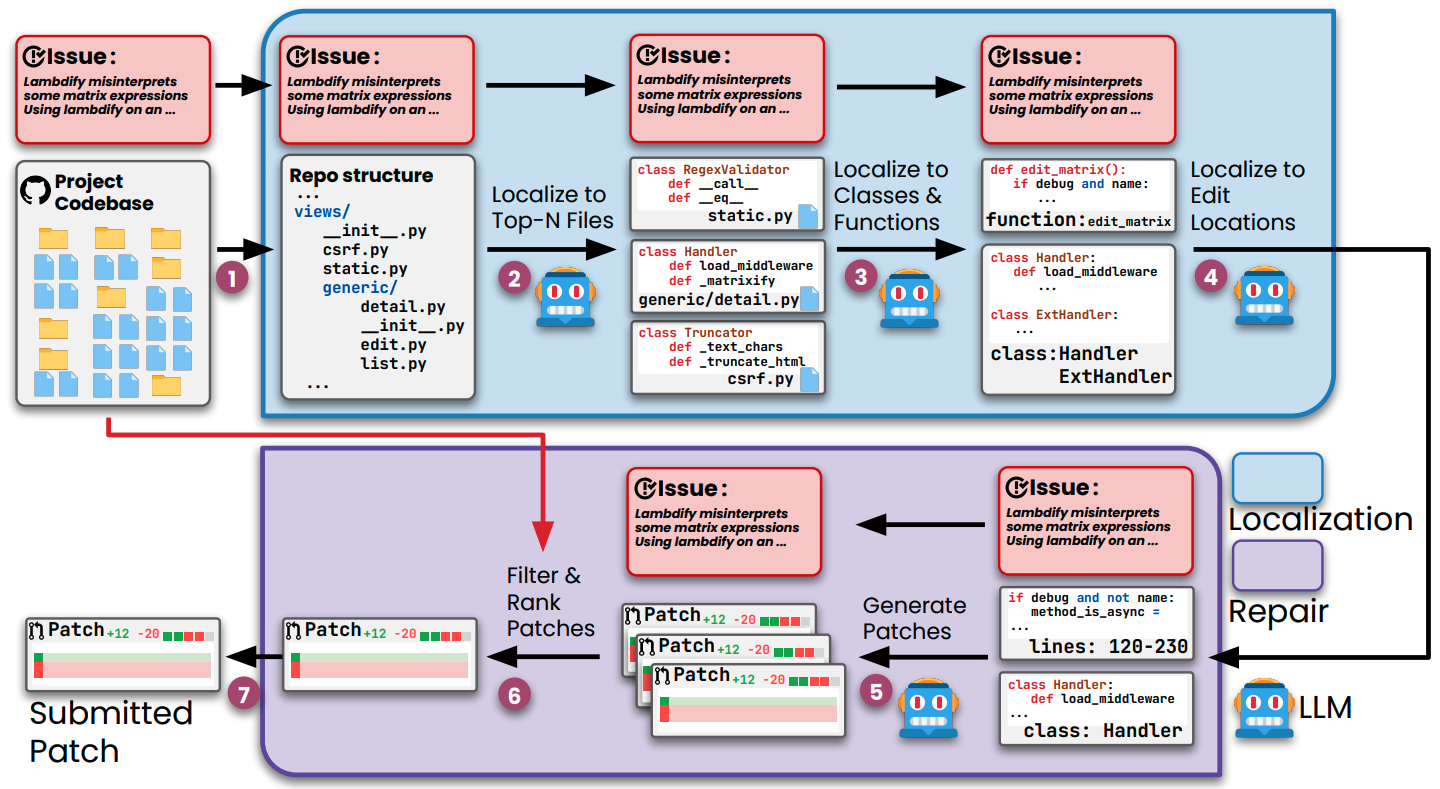}
 \caption{Overview of Agentless, directly taken from their paper \citep{xia2024agentless}.}
\label{fig:agentless}
\end{figure*}

\subsection{Agentless Workflow on SWE-Bench}
\label{app:agentless workflow}

Figure \ref{fig:agentless} is the overview of Agentless taken from their paper \citep{xia2024agentless}.

\end{document}